\documentclass[letterpaper]{article} 
\usepackage{aaai24}  
\usepackage{times}  
\usepackage{helvet}  
\usepackage{courier}  
\usepackage[hyphens]{url}  
\usepackage{graphicx} 
\usepackage{booktabs}
\usepackage{amsmath}
\usepackage[super]{nth}
\usepackage{natbib}  
\usepackage{caption} 
\usepackage{algorithm}
\usepackage{algorithmic}

\usepackage{newfloat}
\usepackage{listings}
\DeclareCaptionStyle{ruled}{labelfont=normalfont,labelsep=colon,strut=off} 
\floatstyle{ruled}
\newfloat{listing}{tb}{lst}{}
\floatname{listing}{Listing}
\title{SkipViT: Speeding Up Vision Transformers\\ with a Token-Level Skip Connection}
\author{
    \textbf{Foozhan Ataiefard, Walid Ahmed, Habib Hajimolahoseini,} \\ \textbf{ Saina Asani, Farnoosh Javadi, Mohammad Hassanpour,} \\ \textbf{ Omar Mohamed Awad, Austin Wen, Kangling Liu, Yang Liu}\\
}
\affiliations{
    Ascend Team, Toronto Research Center, Huawei Technologies \\%
foozhan.ataiefard@huawei.com
}

\usepackage{bibentry}

\begin{document}

\maketitle

\begin{figure*}[t]
  \centering
    \includegraphics[width=.6\textwidth]{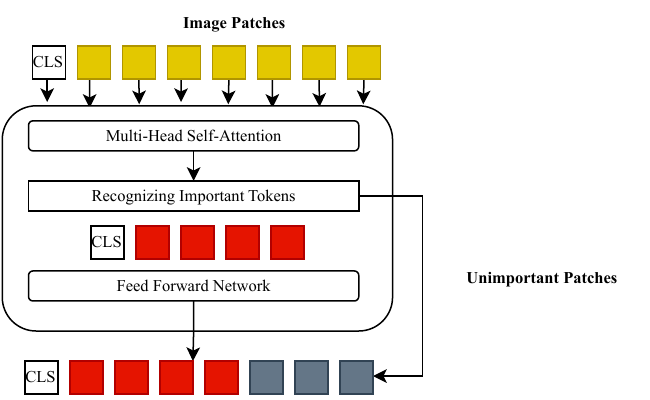}
  \caption{Overview of the SkipViT attention block where the unimportant image patches.}
  \label{fig1}
\end{figure*}

\begin{abstract}
Vision transformers are known to be more computationally and data-intensive than CNN models. These transformer models such as ViT, require all the input image tokens to learn the relationship among them. However, many of these tokens are not informative and may contain irrelevant information such as unrelated background or unimportant scenery. These tokens are overlooked by the multi-head self-attention (MHSA), resulting in many redundant and unnecessary computations in MHSA and the feed-forward network (FFN). In this work, we propose a method to optimize the amount of unnecessary interactions between unimportant tokens by separating and sending them through a different low-cost computational path. Our method does not add any parameters to the ViT model and aims to find the best trade-off between training throughput and achieving a 0\% loss in the Top-1 accuracy of the final model. Our experimental results on training ViT-small from scratch show that SkipViT is capable of effectively dropping 55\% of the tokens while gaining 13.23\% training throughput and maintaining classification accuracy at the level of the baseline model on Huawei Ascend910A.
\end{abstract}

\section{Introduction}
In recent years, Transformer architectures have not only advanced rapidly but have also begun to dominate a myriad of fields \cite{vaswani2023attention}. They have demonstrated state-of-the-art results in tasks such as computer vision \cite{dosovitskiy2020image, hajimolahoseini2023methods}, natural language processing \cite{devlin2018bert}, sequence classification \cite{ataiefard2023graybox}, trading \cite{9618780}, etc. These class of deep models are constructed by stacking multiple transformer blocks which employ the attention mechanism to extract the relation between input values also called input tokens and to generate output tokens by computing the weighted average of input tokens. To have powerful transformers, parallel attention modules also called multi-head attentions are used to capture different relations among tokens. Transformer-based models have achieved more accurate results compared to their competitors such as convolutional neural networks since they can scale up more effectively when dealing with large datasets. This substantial increase in the number of parameters resulted in several issues such as long training time.
\\\\
During the recent years several studies have been conducted to speed up the training process. At the layer level, a key approach is layer freezing, which omits updating frozen parameters and results in training acceleration. For instance, Low-Rank Adaptation (LoRA) and it's variations introduced by \cite{hu2021lora,ahmed2023speeding,hajimolahoseini2021compressing} , which freezes pre-trained weights and inserts two lightweight low-rank matrices per each frozen layer to gain speed up during fine-tuning. \cite{dettmers2023qlora} extended this approach, called Quantized Low-Rank Adaptation (QLoRA), by quantizing the weights to achieve more efficiency. The layer freezing approach is mainly applicable during fine-tuning when a checkpoint of the pre-trained weights is available. 
\\\\
At the attention module level, a novel approach to reduce complexity is to share and merge attention matrices. \cite{shazeer2019fast} proposed Multi Query Attention (MQA), which uses only one single key-value head across all attention heads. Although MQA could achieve fast inference decoding, its harsh merging resulted in a lower quality model. \cite{ainslie2023gqa} extended the idea by introducing GQA, which groups heads and shares a single key head and value head per subgroup. MQA and GQA methods aim at faster inference of large language models. Recent research has suggested that optimizing the number of heads in a transformer architecture can further improve the performance of transformer models \cite{javadi2023gqkva}.
\\\\
In the token level, several methods have been introduced to reduce the sequence length by detecting and dropping unimportant tokens. For example, \cite{rao2023dynamic} proposed DynamicVIT which uses a trainable prediction module to progressively find and mask the less informative tokens. \cite{yao2022randomltd}, proposed Random-LTD which randomly drops tokens in each transformer layer except the first and last layers. The random selection in this approach can cause loss of data from important tokens, therefor they are returned after each layer. That work benefits from some customized implementations to achieve training speed up.\cite{liang2022patches} presented EVIT that uses an importance score-based metric, i.e. largest attentions scores from class tokens, to keep and drop tokens. EVIT computes the average of attentiveness value across all heads, keeps the K tokens with largest attention values and fuses inattentive tokens into a single token.
\\\\
While an attention-based metric can successfully identify informative tokens, based on our observation, replacing the discarded tokens with the weighted average of them as a single fused token did not show a significant impact on keeping accuracy. In particular, we tried EVIT for training VIT-small without any teacher-student method on HUAWEI AI ASCEND device but could not gain training speed up while preserving accuracy. To overcome the computation cost of EVIT and achieve training speed up, we increased the drop ratio, which resulted in a dramatic drop in accuracy.
\\\\
In this paper, we propose SkipViT, a fast training method for VIT that uses an attention score-based approach. This method reduces the number of tokens by dropping less important ones from an image and leverages from a residual connection. This connection enables the ViT to reutilize the dropped tokens in later layers to compensate for loss in the image data.  Finally, we discuss the efficiency of our method and the experimental results that lead us to the final architecture of SkipViT. 



\begin{table*}[t]
\centering
\setlength{\tabcolsep}{3pt}
  
  \centering
  \begin{tabular}{ccccc}
  
    \toprule
     \textbf{Dropping layers} &  \textbf{Drop ratio}    & \textbf{Skip connection} & \textbf{Throughput (Speedup)} & \textbf{Acc. Top-1(\%)}\\
    \toprule
    ViT-small & --  & -- &  4,503 & 70.17  \\
    \midrule
    6  & 45\%  & fused token & 4,963(+10.23\%) & 68.39(-1.78)\\
    \midrule
     4,7  & 30\%,30\%  & 10 & 5,092(+13.1\%) & 69.73(-0.44)   \\
      4,7   & 30\%,30\% & 11 & \textbf{5,227(+16.09\%)} & 69.4(-0.77)   \\
       6,8  & 35\%,35\% & 10 & 4,711(+4.62\%) & 70.53(+0.36)\\
       6,8  & 35\%,35\% & 11 & 4,838(+7.45\%) & 70.41(+0.24)\\
      6   & 45\% & 11 & 4,944(+9.8\%) & \textbf{70.64(+0.47)}\\
       6  & 50\% & 11 & 5,021(+11.51\%) & 70.27(+0.1)\\
       6  & 55\% & 11 & 5,098(+13.23\%) & 70.16(-0.01)\\
    \bottomrule
  \end{tabular}
    \caption{Performance comparison of various skip connection layers, one and two stage token dropping with different Token dropping ratios. Dropping layers shows in which transformer layer of ViT-small we discarded the tokens. Throughput is measured by the number of samples processed per second. The model using fused token does not use a skip connection and replaces the discarded token with a fused token. Metrics highlighted in bold represent the best results.}
    \label{table1}
\end{table*}

\section{Method}
Our method builds on top the same architecture of the ViT with 12 transformer layers and aims to improve the performance of it while keeping the same accuracy. First, we look into the multi-head attention mechanism architecture and then, describe how the proposed method is applied to the attention components.

\subsection{Attention Score Overview}
Multi-Head Attention (MHA), a crucial component in Transformer models \cite{vaswani2017attention}, is designed to capture diverse aspects of the input data. Each MHA unit consists of multiple attention heads, denoted by $h$ with each head focusing on learning different features. The token inputs $x$ to the attention layer are transformed into three distinct matrices: queries $Q$, keys $K$, and values $V$. These transformations are achieved through a linear transformation. 

The attention mechanism in each head is computed as:
\begin{equation}
\label{att}
    \text{Attention}(Q, K, V) = \text{Softmax}\left(\frac{QK^T}{\sqrt{d}}\right)V
\end{equation}

where $d$ represents the dimensionality of the key (and query) vectors.

Here, $QK^T$ is the dot-product of queries and keys, and $\sqrt d$ is the scaling factor to avoid large values in the dot-product attention. In this paper we refer to the resulting matrix of $Softmax(QK^t/\sqrt{d})$ as the attention scores.



\subsection{Identifying Important Patches}
In ViT models, the first step is to execute tokenization by splitting an input image into patches and transforming each into a token embedding using a  convolutional layer. In the final Transformer layer of ViT, the output corresponding to the \texttt{[CLS]} token is commonly utilized. For tasks such as object detection, this output is attached to a classification head, emphasizing its significance in the overall mechanism. This also means that we can employ the attention scores corresponding to this token to detect important patches of an image \cite{liang2022patches}. 
\\\\
The attention scores consist of a $(n+1) \times (n+1)$ matrix where $n$ is the number of input tokens to the attention unit. Based on the attention Eq. \ref{att}, we can say that each row $i$ in the attention score matrix are coefficients by which other tokens will attend in forming the new $i$ token at the attention unit output.
\\\\
Therefore, the values in the first row of the attention scores indicate how much other tokens contribute to forming the new \texttt{[CLS]} token before it is fed into the MLP layer of the ViT. Since ViT-small has 6 attention heads, we first average the attention scores across the head dimension. We then use these average values to determine which token contributes more significantly to determining the correct class for an image.

\subsection{Skip Connection For Tokens}
By dropping 45\% of the tokens from the \nth{6} transformer layer of ViT-small and adding a single fused token, which incorporates a weighted average of the removed tokens, our model was unable to maintain baseline accuracy while gaining throughput, as shown in Table \ref{table1}. 
\\\\
Non-essential patches in an image, like the background or surrounding regions of an object, often contain minimal information. These areas can typically be ignored without significant impact. Completely disregarding non-essential patches such as the surroundings of an object can dramatically reduce the performance of the ViT, while these areas can marginally guide the model and contribute to the final image classification. 
\\\\
We propose the use of a skip connection for the tokens that would otherwise be discarded. This approach selectively excludes these tokens from contributing to certain transformer layers within the model, while still incorporating them in the final layers. Returning the dropped tokens to their original position among other tokens reduces the impact of token dropping on final classification accuracy of the model.  


\section{Experiments}
We performed all of our experiments using ViT as our baseline architecture. For all of the experiments we used hyper parameters presented in Table \ref{tablep} to train the models from scratch. We trained all of the models using the ImageNet1K dataset \cite{5206848} with resolution 224 using the Ascend version of AdamW \cite{DBLP:journals/corr/abs-1711-05101}. We then evaluated the models on the test set of 50,000 images for classification. The metric used to report accuracy is Top-1(\%) and samples per second(FPS) is used to report training throughput of the models. All of the experiments in this paper are conducted using a cluster of $8\times$ Ascend 910A Devices with 32GB of memory. We report our experimental results for the best token dropping strategy in Table \ref{table1}. 

The results of our study indicate that the token fusion approach adopted from previous works \cite{liang2022patches} is not sufficient for our original model and pre-training setup to maintain the final top-1 accuracy without using more advanced architectures such as \cite{touvron2021training}, other data efficient methods\cite{hajimolahoseini2023swiftlearn, li2021short} or increased compute.

\begin{table}[t]
\centering
\begin{tabular}{cc}
\toprule
    \textbf{Parameter} & \textbf{value} \\
    \bottomrule
    Batch size & 288\\
    Epochs & 100 \\
    Weight Decay & 0.5 \\
    Learning Rate & 1e-3\\
    Warmup LR & 1e-6 \\
    Mixup & 0.1 \\
\bottomrule
\end{tabular}
\caption{Original Model Training Parameters}
\label{tablep}
\end{table}

\begin{table*}
\centering

  \begin{tabular}{cccc}
    \toprule
     \textbf{Dropping layers} &  \textbf{Drop ratio}    & \textbf{Warm-up epochs} & \textbf{Acc. Top-1(\%)}\\
    \midrule
     4,7  & 30\%  & 0 & 67.18   \\
      4,7   & 30\% & 15 & 69.73  \\
    \bottomrule
  \end{tabular}
    \caption{Results of training ViT with and without a warm-up ratio before applying 30\% token dropping at layers 4 and 7 with skip connection to layer 10.}
    \label{table2}
\end{table*}

\subsection{Determining Optimal Layers and Ratios for Token Dropping}
We experimented with two strategies for dropping the tokens. A single and a two stage token dropping (i.e., drop in one or two layers) strategy to find the best trade-off between training performance and final accuracy of the ViT model between these methods. A summary of our experimental results are presented in Table \ref{table1}.
\\\\
In both of the dropping methods we were able to see a relative speedup with limited to no loss in the validation accuracy. In single layer token dropping method, our best method with dropping 55\% of the tokens at layer 6 with skip connection to layer 11 reaches 0.01\% accuracy drop while gaining 13.23\% throughput. Using two stage token dropping approach and drop ratio of 30\% for layers 4 and 7 with skip connection to layer 11, our fastest achieved 16.09\% more FPS and reaching 69.4\% classification accuracy which outperforms the token fusion technique.

\subsection{Finding The Optimal Skip Connection}
To prevent from any degradation in Top-1 accuracy of the ViT model we reuse the dropped tokens in the future layers. Based on our findings, there is a trade-off between the FPS improvement and accuracy degradation depending on the transformer layer that the tokens are returned. Table \ref{table1} indicates that delaying the skip connection by even 1 block can cause a substantial decrease in the accuracy metric. Dropping 30\% of the tokens at layers 4 and 7 and returning them to the sequence at \nth{10} layer compared to returning at \nth{11} layer, achieves 2.99\% higher FPS while loosing 0.33\% accuracy.

\subsection{Effect of Warm-up On Patch Detection Quality}
 In Table \ref{table2}, the results indicate that for a similar dropping ratio (30\%) in the same layers (4 and 7), ViT reaches 2.55\% higher accuracy when first 15 epochs are used as warm-up period before token dropping is applied. Based on the results we can draw the conclusion that the warm-up epochs are a essential part of our token dropping strategy which helps the model to select a more informative set of tokens to keep. Since we aim to train ViT models from the beginning, initially, the attention scores for tokens are derived using randomly initialized weights. As the model gradually learns the global relationships between different tokens after the first few epochs of training, the weights of the attention block become more efficient in detecting important tokens.

\section{Conclusions}
Training large Transformer models from scratch require a huge amount of computation and time. In this paper, we propose SkipViT, an intuitive and stable framework to effectively reduce the amount of computation required to train ViT-based models. SkipViT takes advantage of the attention scores of the \texttt{[CLS]} token to differentiate the computation graph between important from less informative tokens. Furthermore, Our proposed framework achieves a significant speedup with no loss in the accuracy of the model by adding a skip connection from the dropping block to a future transformer block in ViT. Due to resource constraints we where only able to apply our experiments on the small version of ViT using the Imagenet1K dataset. This method shows promising results on the current setup, However it is limited by the size of the model and dataset and should be extended to the larger versions of ViT. Additionally, a larger dataset could be used to train ViT using SkipViT to measure scalability of this method. Another positive aspect of this approach is the reduced memory footprint of the model, which needs to be examined in the future works.

\bigskip

\bibliography{aaai24}

\end{document}